\documentclass{article}
\usepackage{spconf,amsmath,graphicx}
\usepackage{amssymb}
\usepackage{mathrsfs}
\usepackage{epstopdf}
\usepackage{algorithmic}
\usepackage{algorithm}

\title{Hyperspectral Unmixing with Endmember Variability using Partial Membership Latent Dirichlet Allocation }
\name{Sheng Zou and Alina Zare\thanks{The authors wish to thank the National Geospatial-Intelligence Agency for support of this research under the project entitled ``NIP: Functions of Multiple Instances for Hyperspectral Analysis.''}}
\address{ Electrical and Computer Engineering, University of Missouri\\ Electrical and Computer Engineering, University of Florida}
\begin{document}
%\ninept
%
\maketitle
\begin{abstract}
The application of Partial Membership Latent Dirichlet Allocation (PM-LDA) for hyperspectral endmember estimation and spectral unmixing is presented.   PM-LDA provides a model for a hyperspectral image analysis that accounts for spectral variability and incorporates spatial information through the use of superpixel-based ``documents.'' In our application of PM-LDA, we employ the Normal Compositional Model in which endmembers are represented as Normal distributions to account for spectral variability and proportion vectors are modeled as random variables governed by a Dirichlet distribution.  The use of the Dirichlet distribution enforces positivity and sum-to-one constraints on the proportion values.  Algorithm results on real hyperspectral data indicate that PM-LDA produces endmember distributions that represent the  ground truth classes and their associated variability.
\end{abstract}
\begin{keywords}
partial membership, latent dirichlet allocation, PM-LDA, hyperspectral, unmixing, endmember
\end{keywords}
\section{Introduction}
\label{sec:intro}

% 1 paragraph 
% What is hyperspectral image? \\
% what is hyperspectral unmixing (endmember and proportion estimation)? \\
% What is LMM? \\
A hyperspectral image is a data cube consisting of two spatial and one spectral dimension with hundreds of spectral bands with relatively narrow, contiguous bandwidths \cite{landgrebe2002hyperspectral}. Due to limitations in spatial resolution, often pixels in a hyperspectral image are mixed, containing signatures from multiple materials. Hyperspectral unmixing is the task of decomposing the mixed signature associated with each pixel into the spectral signatures of pure materials (\emph{i.e.}, endmembers) and their corresponding proportion values \cite{keshava2002spectral}. \par
% 1-2 paragraphes
% What is spectral variability (cause, endmember as set, endmember as distribution)? \\
% Why endmember as distribution is preferred? \\
% Current endmember as distribution methods in the area (Eches, S-PCUE, proposed method) \\
The majority of hyperspectral unmixing approaches in the literature assume the linear mixing model in which each data point is a convex combination of endmember signatures, 
\begin{equation}
\mathbf{x}_n = \sum_{k=1}^K p_{kn} \boldsymbol{e}_k +\boldsymbol{\epsilon}_n \quad n = 1, \ldots ,N
\label{eqn:ConvexGeoModel}
\end{equation}
such that $p_{kn}\geq 0$ and $\sum_{k=1}^K p_{kn} = 1$ and where $\mathbf{x}_n$ is a $m\times 1$ vector containing the spectral signature of the $n^{th}$ pixel in a hyperspectral image, $N$ is the number of pixels in the image, $K$ is the number of endmembers (or materials) found in the scene,  $p_{kn}$ is the proportion of endmember $k$ in pixel $n$, and $\mathbf{e}_k$ is the $m\times 1$ vector containing the spectral signature of the $k^{th}$ endmember,  $\boldsymbol{\epsilon}_n$ is an error term, and $m$ is the number of spectral bands of the hyperspectral data  \cite{bioucas2012hyperspectral}.  In the standard linear mixing model, an endmember is regarded as a single point in high dimensional space and, thus, does not represent the spectral variability of the endmember. However, methods that assume the linear mixing model with fixed endmember signatures often suffer in accuracy since the spectral signature of a material can vary due to changes in illumination, environmental and atmospheric conditions as well as the intrinsic variability of materials \cite{zare2014endmember}. 

Methods for hyperspectral unmixing and endmember estimation that address spectral variability have been developed in literature.  There are two prominent categories for unmixing methods that address spectral variability: (1) \emph{endmembers as sets} methods \cite{roberts1998mapping, bateson2000endmember, somers2012automated, castrodad2011learning} and  (2) \emph{endmembers as statistical distributions} methods \cite{zare2014endmember}.  In this paper, we present PM-LDA as a new ``endmembers as statistical distributions'' approach. \par

Under the ``endmembers as statistical distributions'' approach each endmember is modeled as statistical distribution and one sample of that distribution is viewed as a possible variant of the material's spectral signature. The most commonly used distribution to represent endmembers is the Normal distribution. Under the \emph{normal compositional model} \cite{stein2003application}, each pixel is modeled as a convex combination of $K$ samples from $K$ Normal endmember distributions. A number of normal compositional model approaches have been developed \cite{zare2014endmember}. Two prominent NCM approaches include the method presented by Eches et al. \cite{eches2010bayesian}  in which, given the endmember mean values, a Markov Chain Mote Carlo (MCMC) sampler is used to estimate proportions and endmember covariances and the method presented by Zare et al. \cite{zare2013sampling} in which an MCMC sampler is used to estimate endmember means and proportion values given known endmember covariances. \par

\section{Partial Membership Latent Dirichlet Allocation}
\label{sec:pmlda}

% 1/2 -3/4 pages
% LDA (equations)\\
% PM-LDA (equations)

Partial Membership Latent Dirichlet Allocation (PM-LDA) \cite{chen2016partial,chenpartial} is an extension of Latent Dirichlet Allocation topic modeling \cite{blei2003latent} that allows \emph{words} to have partial membership in multiple topics.  The use of partial memberships allows for topic modeling given data sets in which crisp topic assignments (as done by LDA) is insufficient since data points (or words) may straddle multiple topics simultaneously.  

\begin{figure}[h]
\begin{center}
\includegraphics[height=4cm]{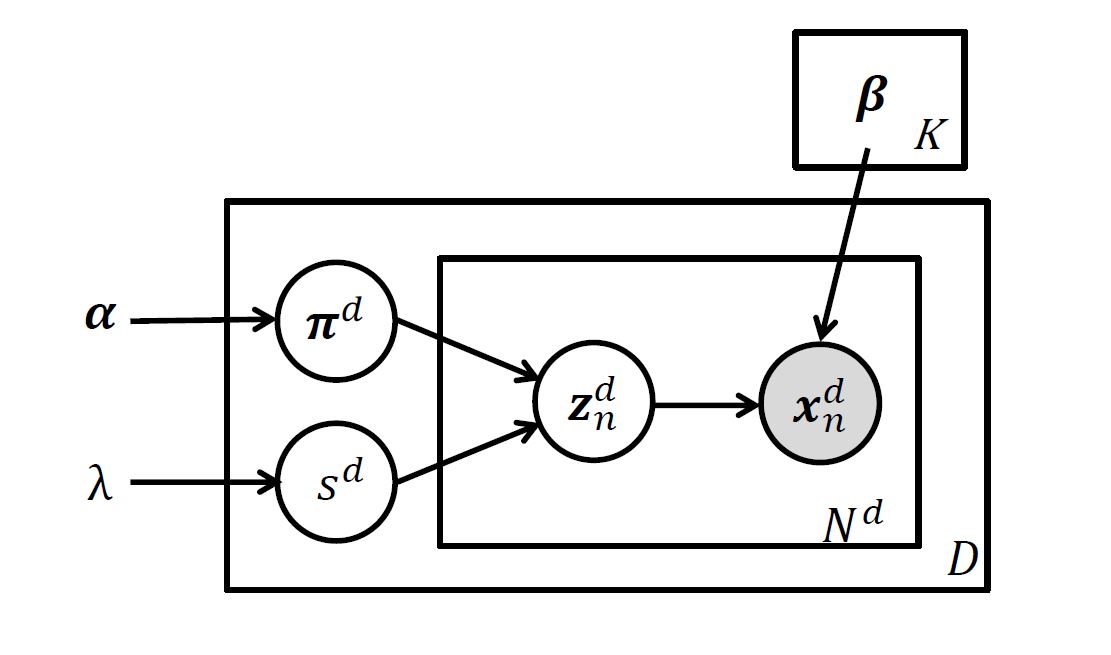}
\caption{Graphical model of PM-LDA}
\label{fig:pmlda}
\end{center}
\end{figure}

The PM-LDA model is a hierarchical Bayesian model in which data is organized at two levels: the \emph{word} level and the \emph{document} level, as illustrated in Fig. \ref{fig:pmlda}.
 In the PM-LDA model, the random variable associated with a data point, $\mathbf{x}$, is assumed to be distributed according to multiple topics with a continuous partial membership in each topic, $\mathbf{z}$. Specifically, the PM-LDA model is
\begin{eqnarray}
p(\boldsymbol{\pi}^d,s^d, \mathbf{z}^d_n,\mathbf{x}^d_n|\boldsymbol{\alpha},\lambda,\boldsymbol{\beta})&=&p(\boldsymbol{\pi}^d|\boldsymbol{\alpha})p(s^d|\lambda)p(\mathbf{z}^d_n|\boldsymbol{\pi}^d,s^d)\nonumber \\
&& \prod_{k=1}^{K}p_k(\mathbf{x}_n^d|\beta_k)^{{z}^d_{nk}} 
%eta
\label{eqn:pmlda}
\end{eqnarray} 
where $\mathbf{x}^d_n$ is the $n$th word in document $d$, $\mathbf{z}^d_n\sim \text{Dir}(\boldsymbol{\boldsymbol{\pi}^ds^d}) $ is the partial membership vector of $\mathbf{x}^d_n$, $\boldsymbol{\pi}^d \sim \text{Dir}(\boldsymbol{\alpha})$ and $s^d \sim \text{exp}(\lambda)$ are the topic proportion vector and  the level of topic mixing in document $d$, respectively. Given hyperparameters $\mathbf{\Psi} = \{\mathbf{\alpha}, \lambda\}$ and the data set (which has been partitioned into documents), $\mathbf{X} = \left\{ \mathbf{X}^1, \mathbf{X}^2, \ldots, \mathbf{X}^D\right\}$, the goal of parameter estimation given the PM-LDA model is to estimate the topic proportion of each document, $\boldsymbol{\pi}^d$, the topic mixing level in each document, $s^d$, the partial memberships of each word in each topic, $\mathbf{z}^d_n$, and the parameters defining the probability distribution of each topic, $\beta_k$. Alg. 1 summarizes a Metropolis-within-Gibbs sampler to perform parameter estimation for PM-LDA \cite{chen2016partial,chenpartial}.
\begin{algorithm}[ht]
\begin{algorithmic}[1]
\REQUIRE{A corpus $\mathbf{D}$, the number of topics $K$, hyperparameters $\mathbf{\Psi} = \{\mathbf{\alpha}, \lambda\}$,  and the number of iterations $T$}
\ENSURE{Collection of all samples: $\boldsymbol{\Pi}^{(t)}, \mathbf{S}^{(t)}, \mathbf{M}^{(t)}$,  $\boldsymbol{\beta}^{(t)}$  
\FOR{$t=1:T$}
\FOR{$d=1:D$}
\STATE \underline{Sample $\boldsymbol{\pi}^d$:} Draw candidate: $\boldsymbol{\pi}^\dagger \sim \text{Dir}(\boldsymbol{\alpha})$ \\
 Accept candidate with probability:\\ $a_{\boldsymbol{\pi}}=\min \left \{ 1, \frac{p(\boldsymbol{\pi}^{\dagger}, s^{(t-1)}, \mathbf{Z}^{(t-1)}, \mathbf{X}|\boldsymbol{\Psi}) p(\boldsymbol{\pi}^{(t-1)}|\boldsymbol{\alpha})}{ p(\boldsymbol{\pi}^{(t-1)}, s^{(t-1)}, \mathbf{Z}^{(t-1)}, \mathbf{X}|\boldsymbol{\Psi}) p(\boldsymbol{\pi}^\dagger|\boldsymbol{\alpha})}\right \}$

\STATE \underline{Sample $s^d$:} Draw candidate: $s^\dagger \sim \text{exp}(\lambda)$\\
Accept candidate with probability:\\
$a_s=\min \left \{ 1, \frac{p(\boldsymbol{\pi}^{(t)}, s^{\dagger}, \mathbf{Z}^{(t-1)}, \mathbf{X}|\boldsymbol{\Psi})p(s^{(t-1)}|\lambda)}{p(\boldsymbol{\pi}^{(t)}, s^{(t-1)}, \mathbf{Z}^{(t-1)}, \mathbf{X}|\boldsymbol{\Psi})p(s^{\dagger}|\lambda)}\right \}$

\FOR{$n=1:N^d$}
\STATE \underline{Sample $\mathbf{z}^d_n$:} Draw candidate: $\mathbf{z}_n^\dagger \sim \text{Dir}(\mathbf{1}_K)$\\
Accept candidate with probability:\\
$a_{\mathbf{z}}=\min \left\{1, \frac{p(\boldsymbol{\pi}^{(t)}, s^{(t)}, \mathbf{z}_n^\dagger, \mathbf{x}_n|\boldsymbol{\Psi})}{p(\boldsymbol{\pi}^{(t)}, s^{(t)}, \mathbf{z}_n^{(t-1)}, \mathbf{x}_n|\boldsymbol{\Psi})}\right\} $
\ENDFOR
\ENDFOR

\FOR{$k=1:K$}
\STATE \underline{Sample $\mu_k$:} Draw proposal: ${\mu}_k^{\dagger}\sim\mathcal{N}(\cdot|{\mu}_{\mathbf{D}},{\Sigma}_{\mathbf{D}})$\\ 
${\mu}_{\mathbf{D}}$ and ${\Sigma}_{\mathbf{D}}$ are  mean and covariance of the data\\
Accept candidate with probability:\\
$a_k=\small{\min\left\{1, \frac{p\left(\boldsymbol{\Pi}^{(t)}, \mathbf{S}^{(t)}, \mathbf{M}^{(t)}, \mathbf{D}|{\mu}_k^{\dagger}\right)\mathcal{N}(\mu_k^{(t-1)}|\mu_\mathbf{D}, \Sigma_\mathbf{D})}{p\left(\boldsymbol{\Pi}^{(t)}, \mathbf{S}^{(t)}, \mathbf{M}^{(t)}, \mathbf{D}|\boldsymbol{\mu}_k^{(t-1)}\right)\mathcal{N}(\mu_k^{\dagger}|\mu_\mathbf{D}, \Sigma_\mathbf{D})} \right\}}$
\ENDFOR
\STATE \underline{Sample covariance matrices $\boldsymbol{\Sigma}= \sigma^2\mathbf{I}$:}\\
Draw candidate from: $\sigma^2 \sim$ Unif$(0,u)$\\ with $u = \frac{1}{2}\left\{\max_{\mathbf{x}_n}d^2(\mathbf{x}_n-{\mu}_{\mathbf{D}})-\min_{\mathbf{x}_n}d^2(\mathbf{x}_n-{\mu}_{\mathbf{D}}) \right\}$\\
Accept candidate with probability: \\
$a_{\boldsymbol{\Sigma}}= \min\left\{1, \frac{p\left(\boldsymbol{\Pi}^{(t)}, \mathbf{S}^{(t)}, \mathbf{M}^{(t)}, \mathbf{D}|\boldsymbol{\Sigma}^\dagger\right)}{p\left(\boldsymbol{\Pi}^{(t)}, \mathbf{S}^{(t)}, \mathbf{M}^{(t)}, \mathbf{D}|\boldsymbol{\Sigma}^{(t-1)}\right)} \right\}.$
\ENDFOR}
\end{algorithmic}
\caption{Metropolis-within-Gibbs Sampling Method for Parameter Estimation}
\label{alg:clustercenter}
\end{algorithm}

\section{Application of PM-LDA for NCM}
\label{sec:relationship}
For application of PM-LDA to hyperspectral NCM unmixing, the hyperspectral scene is first over-segmented into spatially-contiguous superpixels, as shown in Fig. \ref{fig:sub12}.  Each superpixel is assumed to be a \emph{document}.  Given the superpixel segmentation and the assumption that the topic distributions are Gaussian (to assume the NCM), then the parameters of the PM-LDA model can be directly related to parameters of interest in the NCM unmixing model.  Namely, the $K$ topic distributions governed by parameters $\beta_k = \left\{\mu_k, \Sigma_k\right\}$ correspond to the $K$ Gaussian endmember distributions.  The partial membership vector for data point $n$ in document $d$, $\mathbf{z}_n^d$, is the proportion vector associated with the $n^{th}$ data point in the $d^{th}$ superpixel.  The topic proportion vectors for a document, $\boldsymbol{\pi}^d$, correspond to the average proportion vector for a superpixel with the mixing level $s^d$ corresponding to how much each proportion vector in the document is likely to vary from the average proportion vector.  Thus, an entire hyperspectral scene is modeled as a corpus in PM-LDA. 

An advantage of the use of a superpixel segmentation to define documents during unmixing is that it allows us to leverage the expected similarity of the materials found in neighboring pixels.  In other words, spectrally homogeneous neighborhoods are likely to be grouped within a superpixel. Using PM-LDA, all of the pixels in a superpixel are paired with proportion vectors drawn from the same Dirichlet distribution and a shared average proportion vector ($\boldsymbol{\pi}^d$); the variance around that mean is governed by $s^d$.  Larger $s^d$ values correspond to more spatially and spectrally homogeneous superpixels.  

\section{EXPERIMENTS}
\label{sec:experiments}

% 1.5 - 2 pages
%Experiment setup. \\
%Experimental results.
The proposed PM-LDA approach for hyperspectral unmixing was applied to  two hyperspectral images and compared with two previous NCM-based approaches.
\subsection{Experimental setup}

\begin{figure}[h]
\begin{center}
\includegraphics[height=3cm]{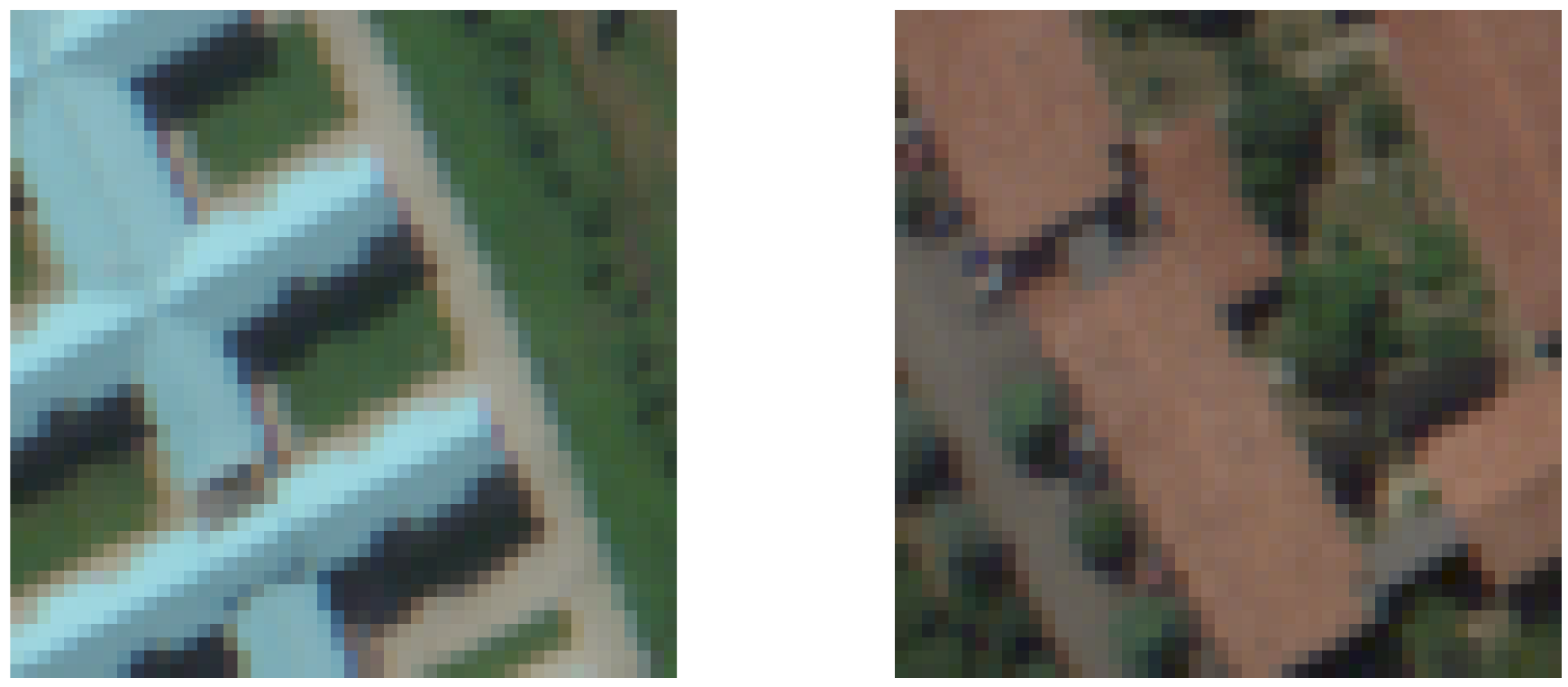}
\caption{University of Pavia sub-images}
\label{fig:sub12orig}
\end{center}
\end{figure}

Two hyperspectral sub-images were extracted from the University of Pavia hyperspectral image data set.  The full hyperspectral scene was collected by the Reflective Optics System Imaging Spectrometer (ROSIS) over an urban area of Pavia in northern Italy on July 8, 2002 and contains 610 by 340 pixels and 103 spectral bands. Two $50\times50$ sub-images were chosen as the test images.  These sub-images are shown in Fig. \ref{fig:sub12orig}.

\begin{figure}[h]
\begin{center}
\includegraphics[height=3cm]{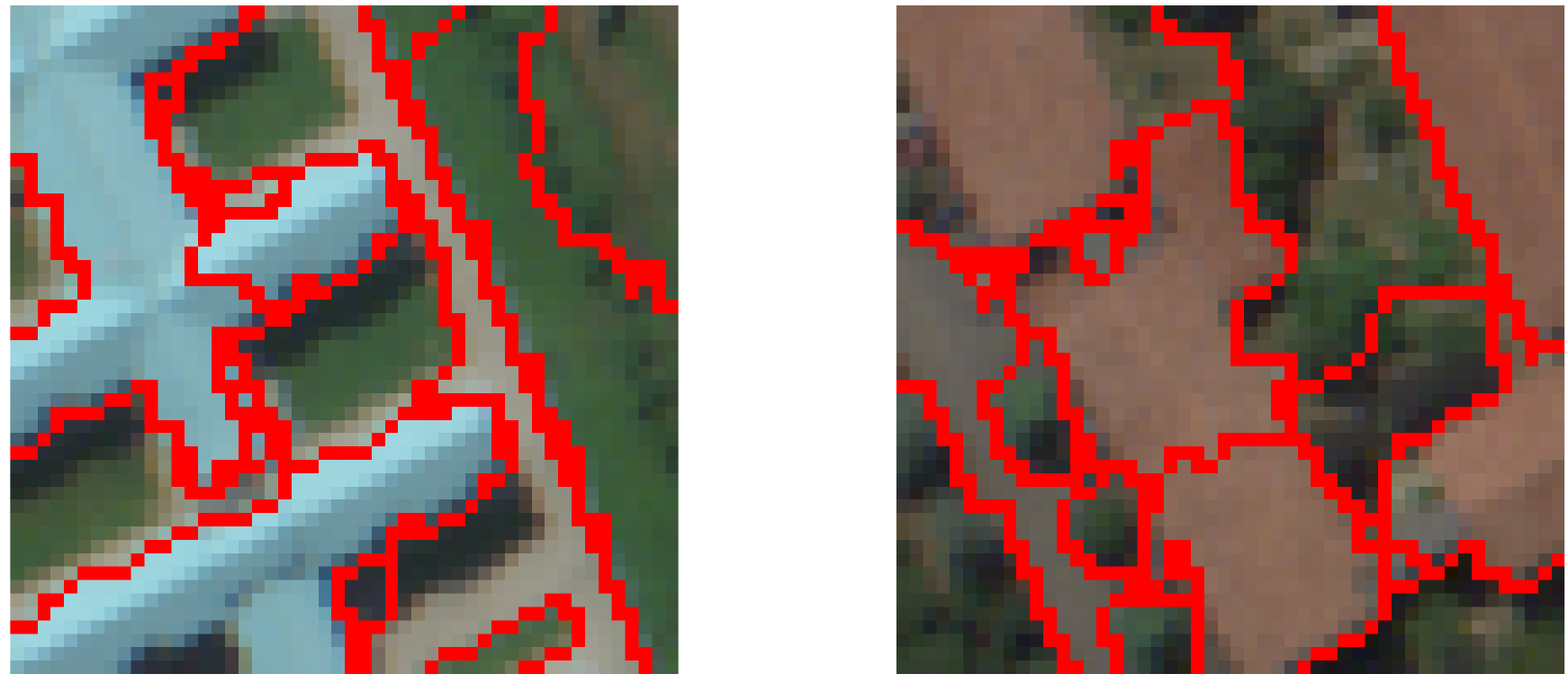}
\caption{University of Pavia dataset segmented sub-images}
\label{fig:sub12}
\end{center}
\end{figure}

The test images are segmented into superpixels using the normalized cuts algorithm presented by Gillis, \emph{et. al.} \cite{gillis2012hyperspectral} that incorporates both spatial and spectral information present in the HSI data.   Fig. \ref{fig:sub12} shows the resulting segmented imagery, each consists of 11 superpixels.  As a pre-processing step, each pixel is normalized to have unit length. The endmember means for the proposed PM-LDA method and the two comparison methods, NCM-Bayes and S-PCUE, \cite{eches2010bayesian, zare2013sampling},  were all initialized with the Vertex Component Analysis (VCA) endmember extraction algorithm \cite{nascimento2005vertex} with same preset number of endmembers.  For PM-LDA and the comparison methods, the parameter settings used in these experiments were selected manually for best performance and are listed as follows: For PM-LDA, $K = 4$ and $3$ for sub-image 1 and 2, respectively, $\lambda = 1$, $\alpha = 5$ and $T = 2000$.  For NCM-Bayes, the Markov chain length was set to 250, the length of the burn-in period was set to 1000, $\delta = 0.001$, and the initial endmember variance was set to $0.001$.  For S-PCUE, the maximum number of endmembers was set to 4 and 3 for sub-image 1 and 2, respectively, the initial endmember variance was set to 0.001, the number of clusters was set to 3 and the maximum number of iterations was set to 10,000. 
%\vspace{-5mm}
%\vspace{-5mm}

\subsection{Experimental results}
\begin{figure}[ht]
\begin{center}
\includegraphics[height=3cm]{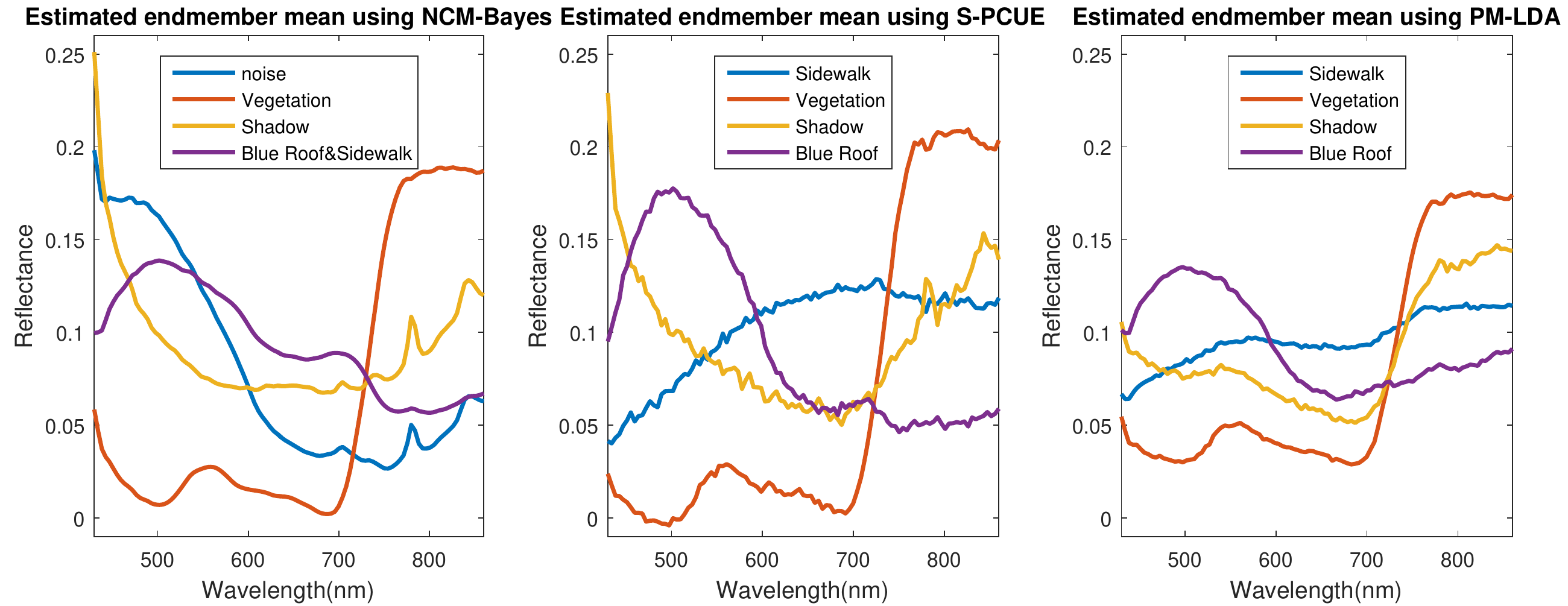}
\caption{Estimated endmember means for sub-image 1 using: (a) NCM-Bayes, (b) S-PCUE, (c) PM-LDA}
\label{fig:e_sub1}
\end{center}
\end{figure}

\begin{figure}[ht]
\begin{center}
\includegraphics[height=5cm]{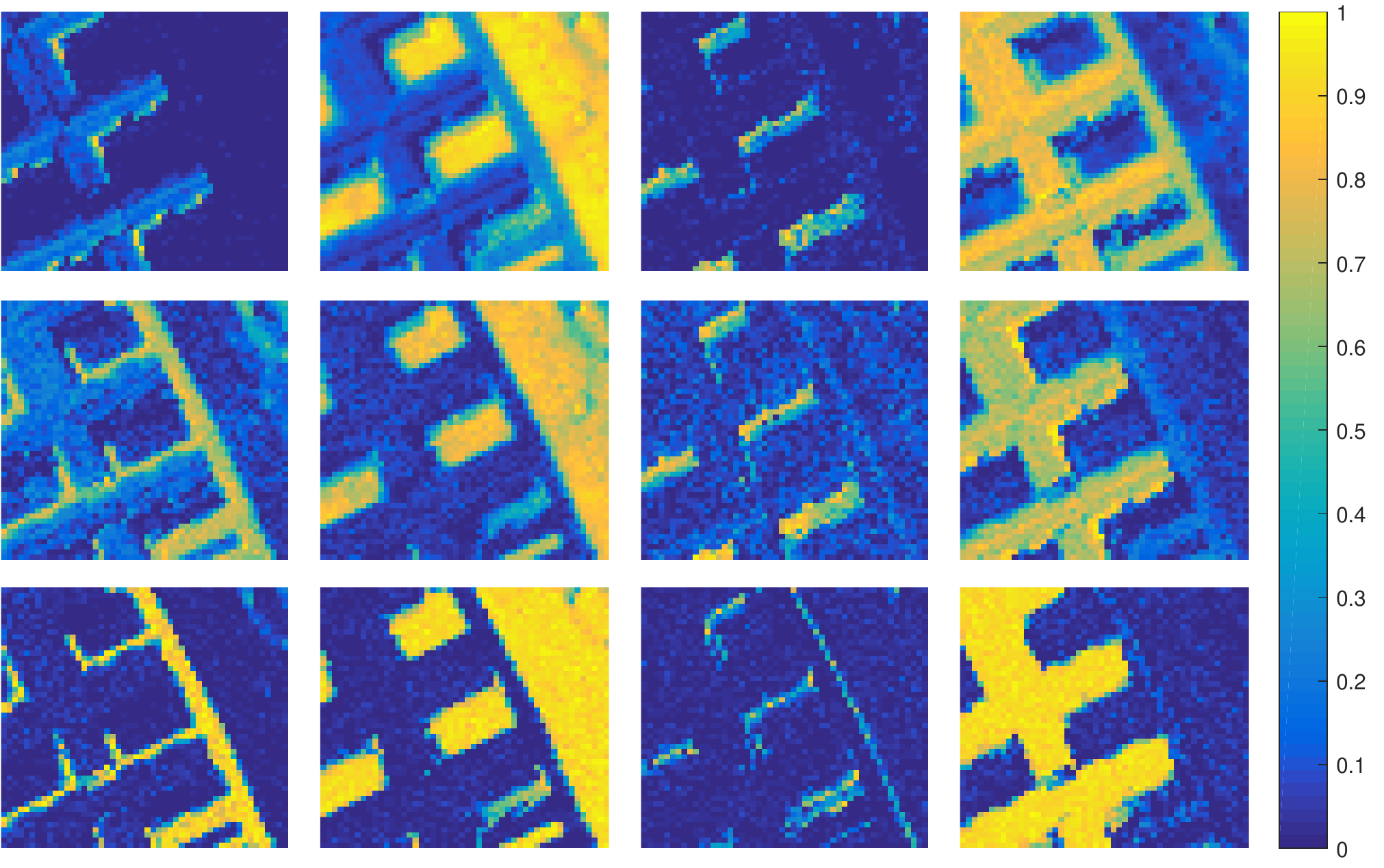}
\caption{Estimated proportion map for sub-image 1: (a) top: NCM-Bayes, (b) middle: S-PCUE, (c) bottom: PM-LDA}
\label{fig:p_sub1}
\end{center}
\end{figure}

The estimated endmember means and proportion maps for sub-image 1 are shown in Fig. \ref{fig:e_sub1} and Fig. \ref{fig:p_sub1}, respectively. In comparison with NCM-Bayes and S-PCUE, as shown in Fig. \ref{fig:p_sub1}, the proportion maps estimated by PM-LDA are qualitatively found to be smooth and the estimated proportion values are high for corresponding pixels dominated by single material and low for others. In contrast, for example, NCM-Bayes is not able to separate blue roof pixels from sidewalk pixels accurately. Qualitatively, S-PCUE appears to favor overly mixed proportion maps (indicating mixing between endmembers) over regions of pure pixels.  \par
\begin{figure}[ht]
\begin{center}
\includegraphics[height=3cm]{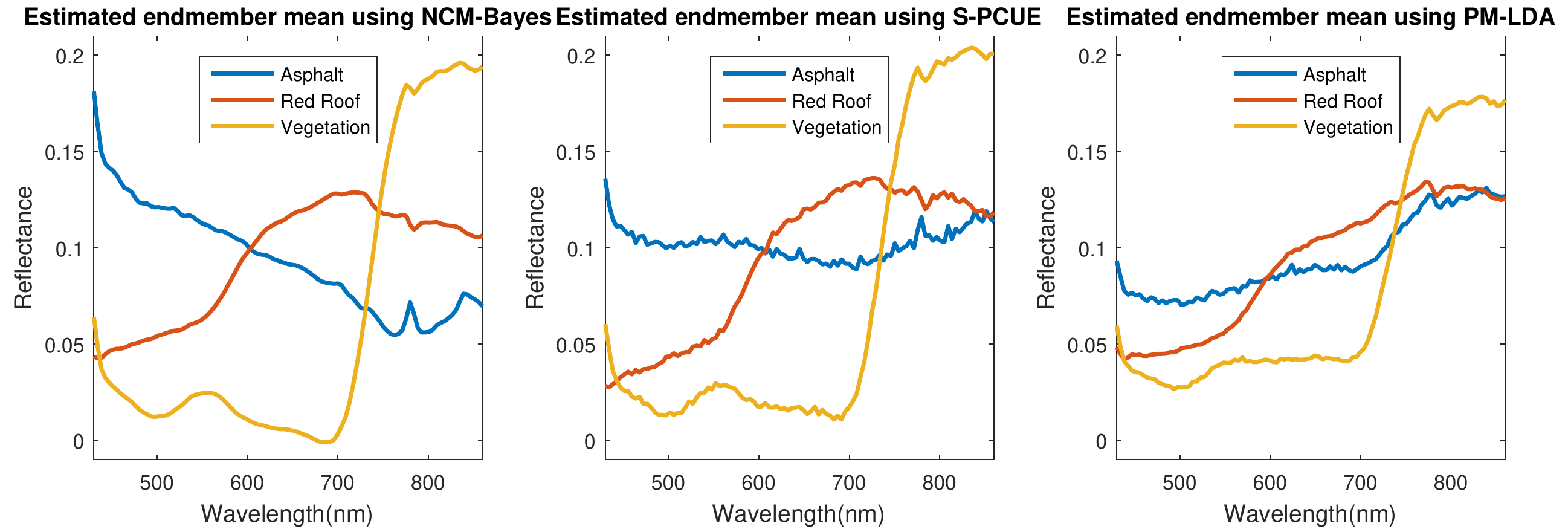}
\caption{Estimated endmember means for sub-image 2 using: (a) NCM-Bayes, (b) S-PCUE, (c) PM-LDA}
\label{fig:e_sub2}
\end{center}
\end{figure}
\begin{figure}[ht]
\begin{center}
\includegraphics[height=6cm]{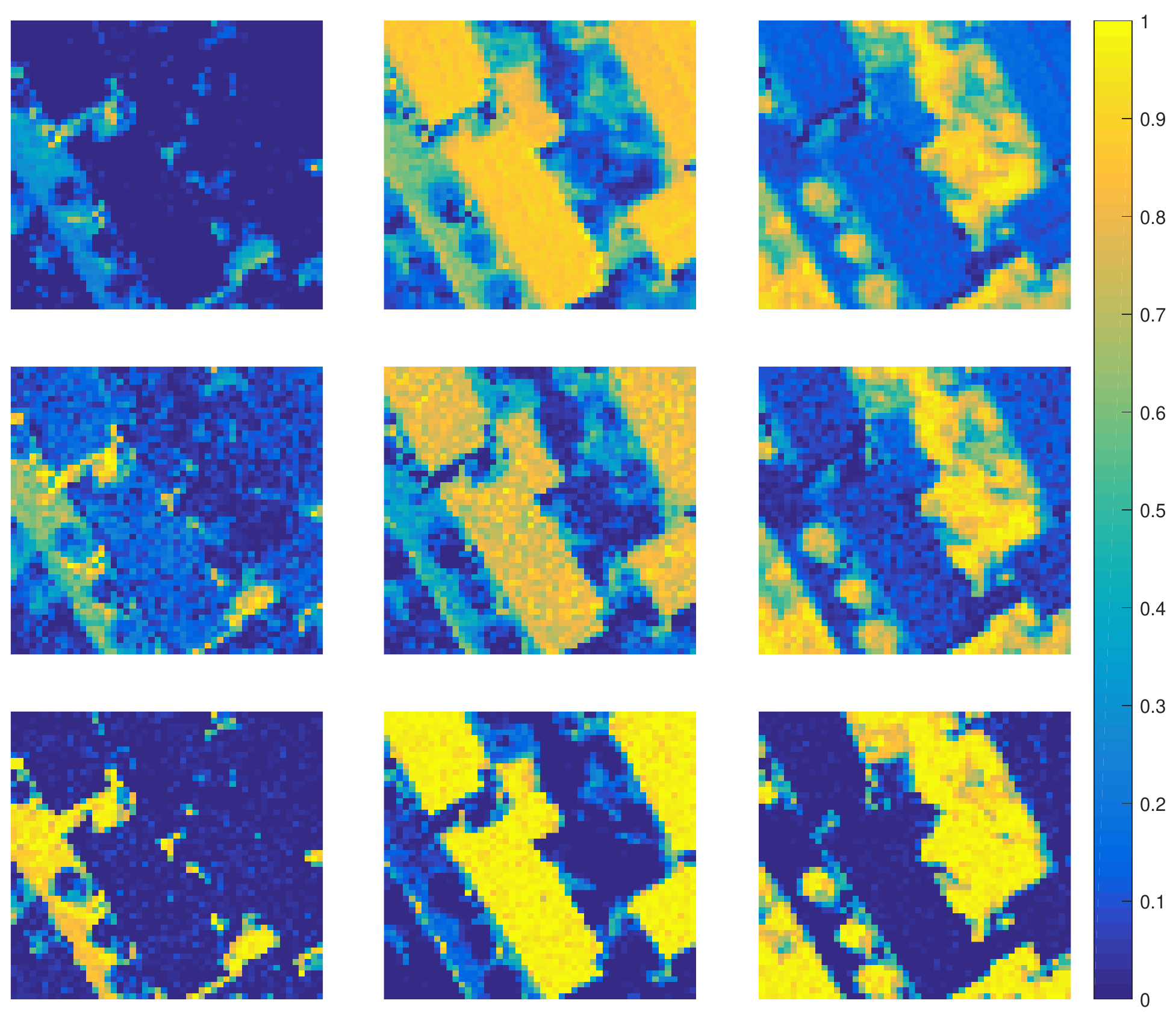}
\caption{Estimated proportion map for sub-image 2: (a) top: NCM-Bayes, (b) middle: S-PCUE, (c) bottom: PM-LDA}
\label{fig:p_sub2}
\end{center}
\end{figure}

Similar results are obtained on sub-image 2.   The estimated endmember means and proportion maps for sub-image 2 are shown in Fig. \ref{fig:e_sub2} and Fig. \ref{fig:p_sub2}, respectively. PM-LDA, in comparison to S-PCUE and NCM-Bayes, provides the only results that estimates both visually accurate and smooth proportion maps. \par

For quantitative evaluation of three approaches, we used two evaluation metrics.  The first evaluation metric was proportion entropy as defined below, 
\begin{equation} \label{eq8}
H\left(\mathbf{P}\right) = -\sum_{n=1}^{N} \sum_{k=1}^{K} p_{nk} \ln p_{nk} 
\end{equation}
where $p_{nk}$ is the proportion value for the $n^{th}$ pixel and $k^{th}$ endmember, $N$ is the number of pixels and $K$ is the number of endmembers.
Proportion entropy is a meaningful evaluation metric since most HSI data points in the test scenes contain only one or, at most, two endmembers.  Thus, accurate proportion vectors for each pixel should have low overall entropy indicating that only one or two proportion values are significant.
The second evaluation metric is the NCM log-likelihood over all pixels in the sub-images. The NCM log-likelihood provides a measure of the overall fit between the test hyperspectral data and the  endmember mean values and covariance values under the NCM model, 
\begin{small}
\begin{equation} \label{eq9}
%f\left(\mathbf{X}|\mathbf{E},\mathbf{P},\mathbf{\Sigma}\right) = \sum_{n=1}^{N} \frac{1}{\sqrt{(2\pi)^L |\sum_{k=1}^{K} p_{nk}^{2} \mathbf{\Sigma}_k |}} \text{exp}\left(-\frac{1}{2}(x_n-\sum_{k=1}^{K}p_{nk}\mathbf{\mu}_k)\right)
f\left(\mathbf{X}|\mathbf{E},\mathbf{P},\mathbf{\Sigma}\right) = \sum_{n=1}^{N} \ln \mathscr{N}\left(\mathbf{x}_n \Bigg|\sum_{k=1}^{K}p_{nk}\mathbf{e}_k,\sum_{k=1}^{K}p_{nk}^{2}\mathbf{\Sigma}_k\right)
\end{equation}
\end{small}

The quantitative entropy and log-likelihood metrics are shown  in Table \ref{tab:entropy} and Table \ref{tab:likelihood}, respectively. PM-LDA achieves the competitive performance on both metrics, showing the lowest proportion entropy and highest NCM overall data log-likelihood. 

\begin{table}[!htb]
\centering  \
\begin{tabular}{lccc} 
\hline
Dataset &NCM-Bayes &S-PCUE &PM-LDA\\ \hline  
Sub-image1 &1542 &2023 &\textbf{1041}\\        
Sub-image2 &1411 &1624 &\textbf{715}\\  \hline
\end{tabular}
\caption{Overall proportion map entropy for three methods}
\label{tab:entropy}
\end{table}
\begin{table}[!htb]
\centering  \
\begin{tabular}{lccc} 
\hline
Dataset &NCM-Bayes &S-PCUE &PM-LDA\\ \hline  
Sub-image 1 &$7.11e5$ &$7.19e5$ &$\textbf{7.31e5}$\\        
Sub-image 2 &$\textbf{8.17e5}$ &$7.04e5$ &${8.04e5}$\\  \hline
Sub-images 1\& 2 &$15.28e5$ &$14.23e5$ &$\textbf{15.35e5}$\\  \hline
\end{tabular}
\caption{Overall log-likelihood for three methods}
\label{tab:likelihood}
\end{table}

\section{Summary}
\label{sec:conclusion}

This article presents application of PM-LDA for NCM-based endmember estimation and unmixing problem in hyperspectral imagery. Experiments show that PM-LDA is effective in addressing the spectral variability.  In the experiments shown, PM-LDA was found to outperform other the two NCM methods both qualitatively and quantitatively. In addition, PM-LDA is the only method of the three that is able to estimate endmember means, endmember covariances, and proportion vectors simultaneously. 
%\section{REFERENCES}
%\label{sec:refs}

\bibliographystyle{IEEEbib}
\bibliography{strings,refs}

\end{document}